\documentclass[sigconf,nonacm]{acmart}
\setcopyright{none}
\usepackage{tabularx}
\usepackage{array}
\usepackage{subcaption}
\usepackage{enumitem}
\usepackage{bbm} 
\usepackage{multirow} 
\usepackage{placeins} 
\newcolumntype{Y}{>{\centering\arraybackslash}X}

\AtBeginDocument{%
  }

\setcopyright{acmlicensed}
\copyrightyear{2026}
\acmYear{2026}
\acmDOI{XXXXXXX.XXXXXXX}
\acmConference[KDD '26]{Proceedings of the 32nd ACM SIGKDD Conference on Knowledge Discovery and Data Mining}{August 9--13, 2026}{Jeju, Korea}
\acmISBN{978-1-4503-XXXX-X/2026/08}

\begin{document}

\title{DeepTaxon: An Interpretable Retrieval-Augmented Multimodal Framework for Unified Species Identification and Discovery}

\author{Jiawei Wang$^{1}$, Ming Lei$^{2}$, Yaning Yang$^{3}$, Xinyan Lin$^{2}$, Yuquan Le$^{4}$, Qiwei Ma$^{2}$, Zhiwei Xu$^{5,7}$, Zheqi Lv$^{6}$, Yuchen Ang$^{1}$, Zhe Quan$^{2*}$, Tat-Seng Chua$^{1}$}
\affiliation{%
  \institution{$^{1}$National University of Singapore \quad $^{2}$Hunan University \quad $^{3}$Xiangtan University \quad $^{4}$Hunan Normal University \quad $^{5}$Zhejiang University \quad $^{6}$Cornell University \quad $^{7}$Meituan}
  \country{}
}
\renewcommand{\authorsaddresses}{$^{*}$Corresponding author: quanzhe@hnu.edu.cn}

\renewcommand{\shortauthors}{Wang et al.}

\begin{abstract}
Identifying species in biology among tens of thousands of visually similar taxa while discovering unknown species in open-world environments remains a fundamental challenge in biodiversity research. Current methods treat identification and discovery as separate problems, with classification models assuming closed sets and discovery relying on threshold-based rejection. Here we present DeepTaxon, a retrieval-augmented multimodal framework that unifies species identification and discovery through interpretable reasoning over retrieved visual evidence. Given a query image, DeepTaxon retrieves the top-$k$ candidate species with $n$ exemplar images each from a retrieval index and performs chain-of-thought comparative reasoning. Critically, we redefine discovery as an explicit, retrieval-based decision problem rather than an implicit parametric memory problem. A sample is novel if and only if the retrieval index lacks sufficient evidence for identification, so each retrieval naturally yields a classification or discovery label without manual annotation, thereby providing automatic supervision for both tasks. We train the framework via supervised fine-tuning on synthetic retrieval-augmented data, followed by reinforcement learning on hard samples, converting high-recall retrieval into high-precision decisions that scale to massive taxonomic vocabularies. Extensive experiments on a large-scale in-distribution benchmark and six out-of-distribution datasets demonstrate consistent improvements in both identification and discovery. Ablation studies further reveal effective test-time scaling with candidate count $k$ and exemplar count $n$, strong zero-shot transfer to unseen domains, and consistent performance across retrieval encoders, establishing an interpretable solution for biodiversity research.
\end{abstract}




\maketitle

\begin{figure}[t]
    \centering
    \begin{subfigure}[t]{0.31\linewidth}
        \centering
        \includegraphics[width=\linewidth]{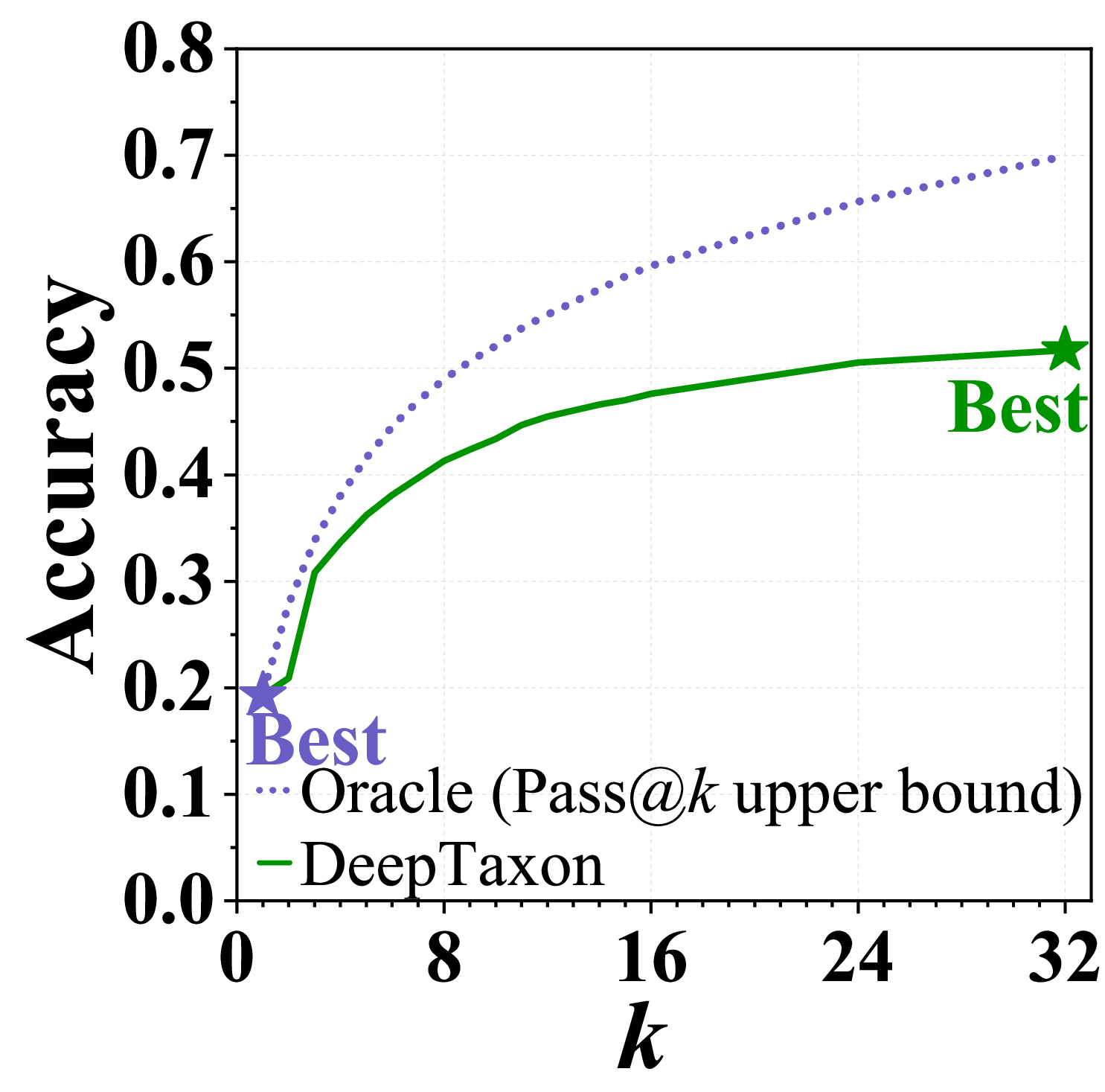}
        \caption{ViT-B}
        \label{fig:passk_vitb}
    \end{subfigure}\hfill
    \begin{subfigure}[t]{0.31\linewidth}
        \centering
        \includegraphics[width=\linewidth]{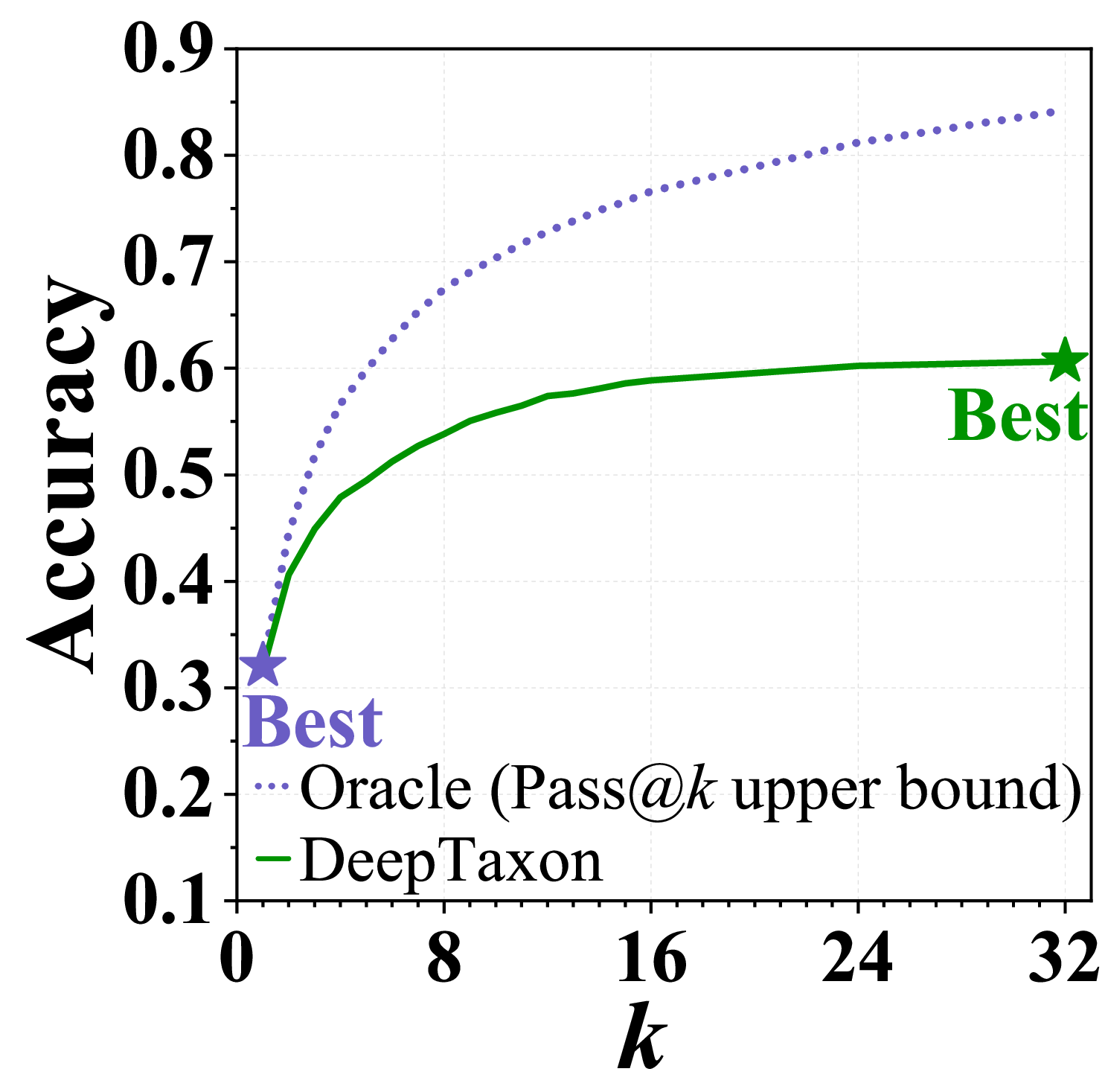}
        \caption{ViT-L}
        \label{fig:passk_vitl}
    \end{subfigure}\hfill
    \begin{subfigure}[t]{0.31\linewidth}
        \centering
        \includegraphics[width=\linewidth]{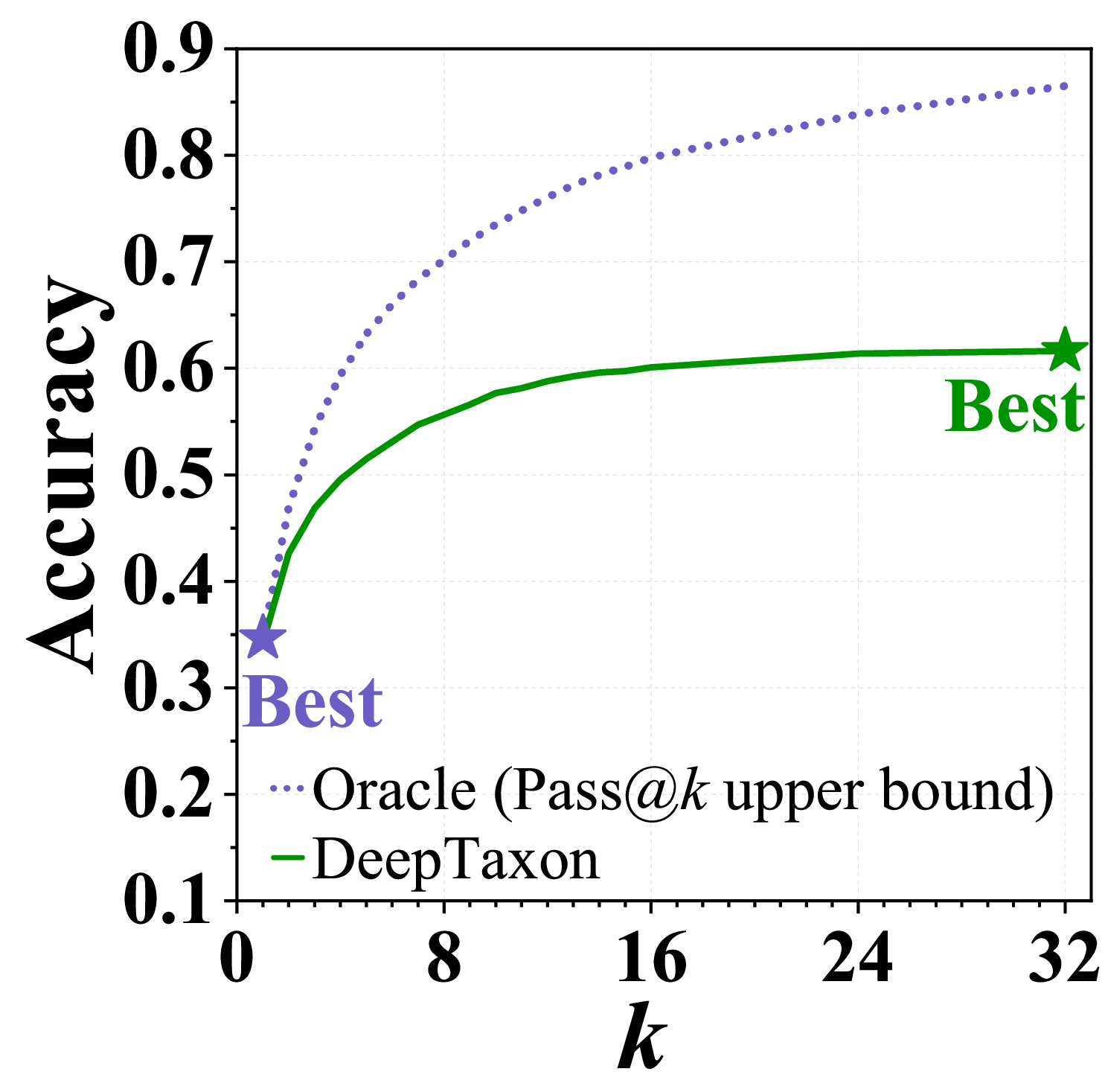}
        \caption{ViT-H}
        \label{fig:passk_vith}
    \end{subfigure}
    \caption{Pass@$k$ curves as a function of retrieved species count $k$ on iNaturalist-10K. Pass@$k$ measures whether the ground-truth species appears among the top-$k$ distinct species retrieved. The star markers denote the best classification accuracy of conventional top-1 retrieval and DeepTaxon, respectively, revealing a significant retrieval-decision gap that DeepTaxon substantially narrows. Detailed numerical values are provided in Appendix~\ref{app:passk_data}.}
    \label{fig:teaserfigure}
\end{figure}
\section{Introduction}
\label{sec:introduction}

\begin{figure*}[t]
    \centering
    \includegraphics[width=0.95\textwidth]{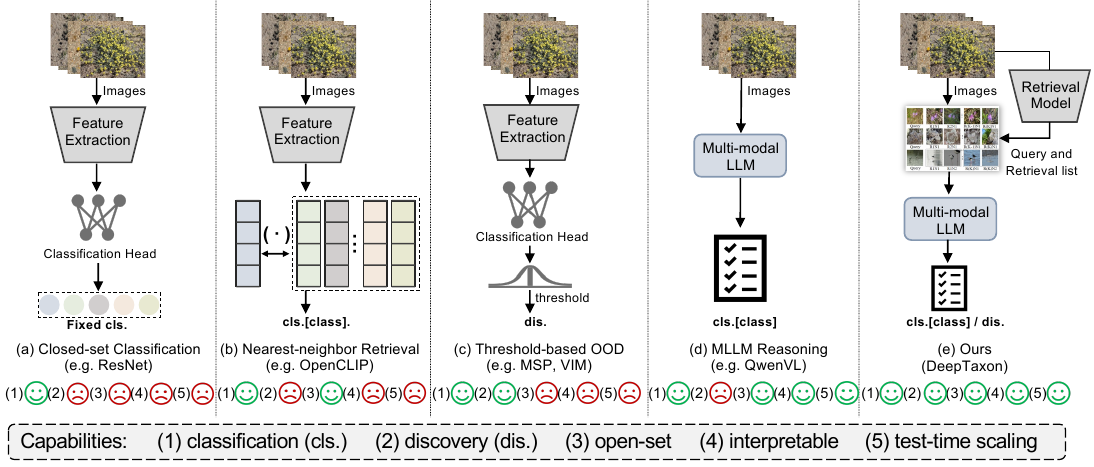}
    \caption{Comparison of five paradigms for species identification and discovery, evaluated across five capabilities: (1) classification, (2) discovery, (3) open-set generalization, (4) interpretability, and (5) test-time scaling. Existing approaches achieve only a subset of these capabilities. DeepTaxon uniquely satisfies all five requirements through retrieval-augmented multimodal reasoning.}
    \label{fig:paradigm}
\end{figure*}

Automated species identification supports biodiversity monitoring~\cite{norouzzadeh2018automatically}, ecological research~\cite{waldchen2018machine}, and citizen science platforms~\cite{vanhorn2018inaturalist,sullivan2014ebird} that handle millions of observations annually, which span deeply hierarchical taxonomies with thousands of leaf-level taxa exhibiting high visual similarity. Building effective systems for these applications requires addressing two complementary challenges: accurately classifying images of known species from such massive and fine-grained label spaces, and detecting novel species that fall outside the training distribution. These dual requirements of classification and discovery are essential for practical deployment but have traditionally been treated as separate problems.

As shown in Figure~\ref{fig:paradigm}, existing approaches each address only a subset of these requirements. Closed-set classifiers~\cite{he2016deep,dosovitskiy2021image} achieve high accuracy on known categories but cannot handle novel species. Nearest-neighbor retrieval with vision encoders~\cite{ilharco2021openclip} enables open-set classification but simply selects the top-1 species without any mechanism to reject queries when no match exists. Out-of-distribution detection methods~\cite{hendrycks2017baseline,wang2022vim} support discovery through confidence thresholding, but create an inherent trade-off, as aggressive thresholds improve discovery at the cost of classification accuracy. Multimodal LLMs~\cite{openai2023gpt4v,bai2023qwen} provide interpretable reasoning but hallucinate without visual grounding to reference images. No existing paradigm unifies all five capabilities simultaneously.

We identify a key insight that motivates our approach. As shown in Figure~\ref{fig:teaserfigure}, when retrieving the top-$k$ species by visual similarity, the probability that the correct species is included (Pass@$k$) frequently exceeds 70\% at moderate $k$, yet top-1 accuracy remains substantially lower. This \textit{retrieval-decision gap} reveals that the bottleneck is not retrieval but decision-making, as rich comparative information in the retrieved candidate set remains unexploited by existing methods.

To bridge this gap, we propose DeepTaxon, a retrieval-augmented multimodal reasoning framework. The key idea is to reformulate both classification and discovery as outcomes of unified reasoning over retrieved visual evidence. This formulation is inherently interpretable, as each decision is grounded in chain-of-thought reasoning and retrieved visual evidence. Critically, we redefine discovery as an explicit, retrieval-based decision problem rather than an implicit parametric memory problem~\cite{hendrycks2017baseline,wang2022vim}. Specifically, a sample is novel if and only if the retrieval index lacks sufficient evidence for identification. This enables automatic data synthesis for both tasks, as each retrieval naturally produces a classification label when the correct species appears among candidates and a discovery label otherwise. Unlike traditional OOD detection that relies on datasets with entirely disjoint and often semantically distant labels, our discovery samples emerge from visually similar candidates, yielding stronger supervision signals. Concretely, given a query image, DeepTaxon retrieves $k$ candidate species with $n$ exemplar images each and performs comparative reasoning over the retrieved references. To learn this ability, we synthesize DeepTaxon-SFT-76k through retrieval-augmented context engineering and apply supervised fine-tuning~\cite{ouyang2022training} to teach structured comparative reasoning. To further strengthen both classification and discovery, we construct DeepTaxon-RL-10k by filtering hard in-distribution samples where the model remains uncertain and augmenting with cross-domain queries that share no species with the retrieval index, then apply Group Relative Policy Optimization (GRPO)~\cite{shao2024deepseekmath} to push accuracy toward the Pass@$k$ ceiling. Our key contributions can be summarized as follows:
\begin{itemize}[nosep,leftmargin=*]
    \item We propose DeepTaxon, an interpretable retrieval-augmented multimodal reasoning framework that reformulates species identification and discovery as a unified decision-making problem, shifting discovery from an implicit parametric memory problem to an explicit retrieval-based decision problem.
    \item We design a retrieval-augmented context engineering pipeline to automatically generate balanced classification and discovery supervision, yielding DeepTaxon-SFT-76k for supervised fine-tuning. We then leverage this environment to filter hard in-distribution and cross-domain samples with the SFT model, and apply GRPO to further strengthen identification accuracy toward the retrieval ceiling and improve discovery reliability.
    \item We conduct extensive experiments on a 10K-class species benchmark and six fine-grained cross-domain datasets, demonstrating consistent improvements over existing paradigms in identification and discovery accuracy, cross-domain generalization, retriever-agnostic performance, and test-time scaling capability. Moreover, we have released our source code to facilitate future research.\footnote{\url{https://anonymous.4open.science/r/DeepTaxon/README.md}}
\end{itemize}

\section{Related Work}
\label{sec:related_work}

\begin{figure*}[t]
    \centering
    \includegraphics[width=\textwidth]{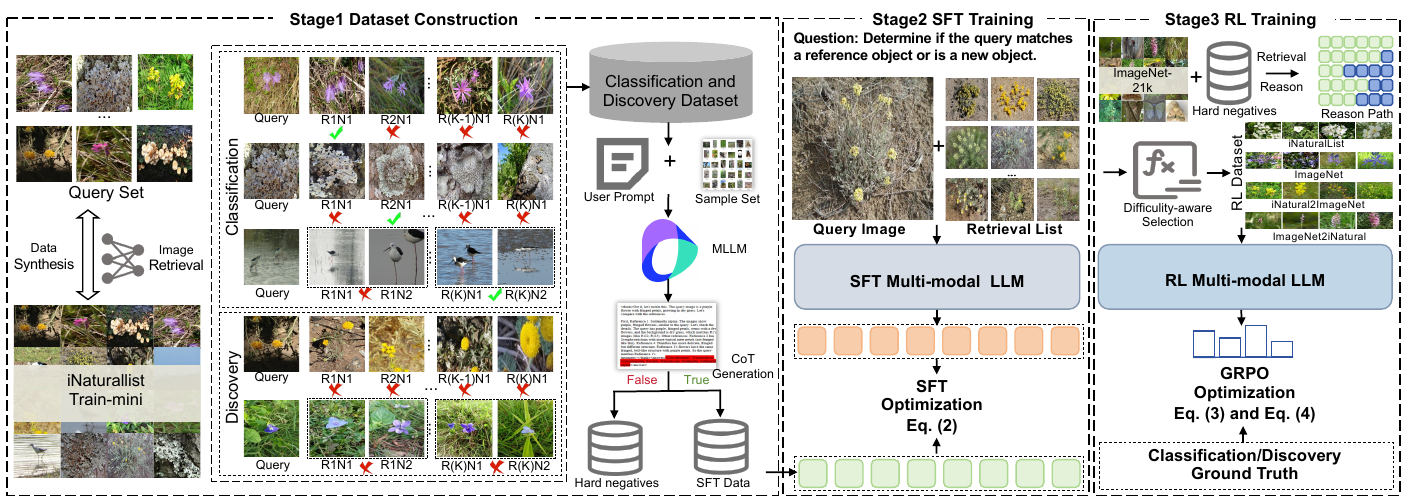}
    \caption{Overview of DeepTaxon. Given a query image, the retrieval module retrieves $k$ candidate species with $n$ exemplars each from the retrieval index. The reasoning module performs comparative analysis and outputs either a taxonomic classification or a discovery signal. Two-stage training with supervised fine-tuning and GRPO optimizes decision quality.}
    \Description{Framework overview diagram showing retrieval, reasoning, and decision components.}
    \label{fig:framework_overview}
\end{figure*}

\subsection{Image Classification and Discovery}

Fine-grained visual recognition distinguishes visually similar subcategories such as species, aircraft variants, or car models. The field has evolved from part-based features and attention mechanisms~\cite{wang2025xr,liu2022focus} to metric learning~\cite{hu2023hierarchical,liao2022graph} and CLIP-based methods~\cite{wu2024building,ilharco2021openclip} that enable zero-shot generalization. Extreme multi-label classification~\cite{chang2020taming,kharbanda2024gandalf} addresses massive label spaces with 100K+ categories, demonstrating scalability beyond traditional benchmarks. However, these approaches assume closed-set classification where all test samples belong to known categories.

Complementary work on out-of-distribution (OOD) detection addresses novel sample identification. Confidence-based methods such as MSP, ODIN, and energy scoring~\cite{yang2022openood,hendrycks2017baseline} threshold classifier outputs to reject unknowns, while feature-based approaches including ViM~\cite{wang2022vim} and ReAct~\cite{sun2021react} detect anomalies through representation geometry. Recent advances address long-tailed distributions~\cite{miao2024ood} and fine-grained open-set scenarios~\cite{lang2024coarse}. Novel Category Discovery~\cite{wang2021progressive} clusters unlabeled data to uncover latent categories offline.

These methods treat classification and discovery as separate pipelines with distinct models, thresholds, and evaluation protocols. The fundamental limitation is that threshold-based rejection relies on a single confidence score to make binary accept/reject decisions, creating an inherent trade-off where improving discovery degrades classification. DeepTaxon eliminates this trade-off through unified retrieval-based reasoning.

\subsection{Retrieval-Augmented Visual Classification}

While the rejection-based methods above address open-set recognition through learned thresholds, retrieval-augmented approaches take a different direction by incorporating external evidence at inference time to improve classification. In NLP, RAG~\cite{lewis2020retrieval} and Atlas~\cite{izacard2023atlas} established the paradigm of grounding model outputs in retrieved knowledge for few-shot generalization, with recent surveys~\cite{fan2024rag} systematically cataloguing retrieval-augmented LLM architectures. In vision, cross-modal neighbor representations~\cite{yi2024leveraging} improve CLIP zero-shot classification by constructing image representations from retrieved textual neighbors, and kNN-based methods~\cite{gui2024knn} enable training-free open-vocabulary recognition by retrieving from an instance embedding database. However, these methods treat retrieved evidence as auxiliary feature augmentation with fixed computation at test time. As shown in Figure~\ref{fig:teaserfigure}, modern retrievers achieve Pass@$k$ exceeding 70\% yet top-1 accuracy remains far lower, revealing a \textit{retrieval-decision gap}. DeepTaxon bridges this gap through interpretable chain-of-thought reasoning over retrieved candidates, where increasing the number of candidates, exemplars, or reasoning tokens all improve performance at inference time.

\subsection{Multimodal Reasoning}

Multimodal LLMs such as GPT-4V~\cite{openai2023gpt4v}, Gemini~\cite{team2023gemini}, LLaVA~\cite{liu2024visual}, and Qwen-VL~\cite{bai2023qwen} exhibit strong cross-modal reasoning. Visual grounding~\cite{peng2024kosmos2} links text to image regions but requires dense annotations unavailable in recognition datasets. Multimodal chain-of-thought~\cite{zhang2023multimodal} separates reasoning from answer inference, motivating our two-stage training. Test-time compute scaling~\cite{snell2024scaling} improves reasoning through chain-of-thought or tree search, while reinforcement learning~\cite{shao2024deepseekmath,wang2025remol} further optimizes reasoning trajectories.

However, MLLMs cannot memorize fine-grained visual features for tens of thousands of taxa, nor reliably generate correct Latin taxonomic names from parametric knowledge alone. DeepTaxon addresses this through unified, interpretable, retrieval-augmented reasoning, grounding each decision in retrieved reference images and taxonomic labels so the model relies on explicit visual comparison rather than parametric memory.

\section{Methodology}
\label{sec:methodology}

\subsection{Problem Formulation}
\label{subsec:problem_formulation}

We formulate species identification and discovery as a unified retrieval-based decision problem. Let $\mathcal{D} = \{(\mathbf{x}_i, y_i)\}_{i=1}^{M}$ denote a retrieval index containing $M$ labeled specimens across $C$ species. Given a query image $\mathbf{x}_q$, the system produces a decision $\hat{d}$ that is either a taxonomic classification $y^* \in \mathcal{Y}$ when evidence supports identification, or a discovery signal when the query likely represents a novel species outside the retrieval index. Both outcomes arise from the same evidence evaluation process, where the system assesses whether retrieved references provide sufficient support for classification or indicate potential novelty.

\subsection{Framework Overview}
\label{subsec:framework_overview}

As illustrated in Figure~\ref{fig:framework_overview}, DeepTaxon consists of three components: (1) a retrieval-augmented environment that retrieves $k$ distinct species with $n$ exemplar images each, whose retrieval outcomes naturally define ground-truth labels for both classification and discovery without manual annotation, (2) a multimodal reasoning module that performs comparative analysis and outputs structured reasoning chains before a decision, and (3) a two-stage training pipeline that first synthesizes retrieval-augmented reasoning data for supervised fine-tuning and then applies GRPO on hard samples to push accuracy toward the retrieval ceiling. The parameters $k$ and $n$ enable test-time scaling~\cite{snell2024scaling}, allowing users to trade computation for accuracy without retraining.

\subsection{Retrieval-Augmented Environment}
\label{subsec:retrieval_environment}

The retrieval module constructs the candidate set from which the reasoning model draws comparative evidence. We employ pretrained OpenCLIP~\cite{ilharco2021openclip} vision encoders as feature extractors and construct a FAISS index~\cite{johnson2019billion} over the retrieval index for efficient nearest neighbor search. Given a query image $\mathbf{x}_q$, we compute its embedding and retrieve candidates based on cosine similarity.

A critical design choice is species-level aggregation. Standard top-$k$ image retrieval often returns multiple images of the same species, wasting the candidate budget on redundant hypotheses and reducing both recall and candidate diversity. We instead retrieve the top-$k$ distinct species by iterating through the similarity-ranked list and selecting species by first appearance. For each selected species $c_j$, we collect its $n$ most similar exemplars to form the reference set $\mathcal{E}_j = \{\mathbf{x}_{j,1}, \ldots, \mathbf{x}_{j,n}\}$. The complete retrieval environment is $\mathcal{R} = \{(c_j, \mathcal{E}_j)\}_{j=1}^{k}$.

We define Pass@$k$ as the probability that the ground-truth species appears among the top-$k$ retrieved distinct species:
\begin{equation}
\text{Pass@}k = \mathbb{E}_{\mathbf{x}_q}\left[\mathbbm{1}\left[y_q \in \mathcal{C}_k(\mathbf{x}_q)\right]\right]
\label{eq:passk}
\end{equation}
where $y_q$ is the true species, $\mathcal{C}_k(\mathbf{x}_q)$ is the set of $k$ retrieved species, and $\mathbbm{1}[\cdot]$ is the indicator function. Pass@$k$ establishes the theoretical ceiling for classification accuracy since correct identification is impossible when the true species is absent from candidates.

A key design insight is that retrieval outcomes naturally determine ground-truth labels for training, providing an automatic and principled definition of discovery. When the true species $y_q \in \mathcal{C}_k$, we label the sample as classification with target $y_q$. When $y_q \notin \mathcal{C}_k$, we label it as discovery. This retrieval-based labeling resolves a key practical limitation of traditional OOD detection, where discovery datasets require test labels to be entirely disjoint from training labels, making data construction costly. In our formulation, any sample whose true species falls outside the candidate list naturally serves as a discovery example, greatly lowering the barrier for automatic data construction and providing strong supervision for both tasks without manual annotation.

\subsection{Multimodal Comparative Reasoning}
\label{subsec:reasoning}

Given the retrieved candidate set $\mathcal{R}$, the reasoning module transforms candidate selection into explicit visual decision-making. The input to the vision-language model consists of the query image $\mathbf{x}_q$ followed by $k$ reference entries, where each entry contains the $n$ exemplar images $\mathcal{E}_j$ and the corresponding taxonomic label of species $c_j$. The model generates structured output consisting of a chain-of-thought reasoning section~\cite{wei2022chain} that articulates step-by-step visual comparisons between the query and each reference, followed by a final decision: either a taxonomic classification selecting one of the $k$ candidates, or a discovery signal indicating that no candidate sufficiently matches the query.

This design achieves interpretability by requiring the model to explain its comparative analysis before committing to a decision. Domain experts can verify predictions by examining which visual features the model considers and how it distinguishes between candidate species. The complete prompt template is provided in Appendix~\ref{app:prompt}.

\subsection{Synthetic Data Construction}
\label{subsec:synthetic_data}

Since retrieval-based labeling (Section~\ref{subsec:retrieval_environment}) automatically provides ground-truth decisions, the remaining challenge is generating the structured reasoning chains that connect retrieved evidence to those decisions. We develop a retrieval-augmented context engineering pipeline to synthesize such reasoning data at scale.

The data construction process partitions training specimens into disjoint query and retrieval pool sets, ensuring queries never appear in their own candidate sets. For each query, we sample $k$ uniformly from $\{4, \ldots, 16\}$ and $n$ from $\{1, 2, 3, 4\}$, then retrieve candidates following the procedure in Section~\ref{subsec:retrieval_environment}. doubao-seed-1.6-vision~\cite{bytedance2025doubao} receives the query and candidates with instructions to perform comparative analysis and output a structured decision. We validate each response against the automatically determined ground truth and retain only samples where the prediction matches, ensuring high annotation quality.

We employ retriever-stratified sampling to construct balanced training data. Different retrieval encoders exhibit different precision profiles. High-precision encoders like ViT-H frequently include the correct species among candidates, yielding classification samples where the model must discriminate among confusers. Lower-precision encoders like ViT-B often exclude the correct species, yielding discovery samples where the model must recognize insufficient evidence. By sampling from encoders with varying precision, we construct a balanced dataset covering both classification and discovery scenarios.

\subsection{Supervised Fine-Tuning}
\label{subsec:sft}

Supervised fine-tuning teaches the model structured reasoning formats and basic comparative strategies. We minimize the negative log-likelihood over complete output sequences:
\begin{equation}
\mathcal{L}_{\text{SFT}} = -\mathbb{E}_{(\mathbf{x}_q, \mathcal{R}, s)} \left[ \sum_{t=1}^{T} \log \pi_\theta(s_t \mid s_{<t}, \mathbf{x}_q, \mathcal{R}) \right]
\end{equation}
where $s = (s_1, \ldots, s_T)$ includes reasoning chains and final decisions. Computing loss over the entire sequence encourages the model to articulate step-by-step analysis before decisions, which is essential for interpretability. The random sampling of $(k, n)$ during training exposes the model to diverse environment configurations, enabling robust generalization at inference time.

\subsection{Reinforcement Learning with GRPO}
\label{subsec:rl}

Supervised fine-tuning data only includes samples that the annotation model solves correctly, creating distribution shift toward easier cases. Challenging samples where visual distinctions are subtle or evidence is ambiguous remain underrepresented. To address this limitation and push performance toward the Pass@$k$ ceiling, we employ reinforcement learning on hard samples.

We construct the hard sample dataset by retaining queries where the annotation model fails during synthetic data construction, representing the frontier of task difficulty. We augment with cross-domain samples using ImageNet-21k~\cite{ridnik2021imagenet21k} to strengthen discovery learning. Specifically, we use ImageNet-21k queries against the iNaturalist index and vice versa. Since these domains share no overlapping categories, cross-domain samples provide strong supervision for discovery prediction.

To balance training across different retrieval complexities, we sample training queries with $k$-proportional weighting: the number of samples at each $k \in \{4, \ldots, 16\}$ is proportional to $k$ itself (ratio $4{:}5{:}6{:}\cdots{:}16$), allocating more training signal to larger candidate sets where discrimination is more challenging. For each query, we generate $G=8$ rollouts from the SFT model and count the number of correct predictions. We apply asymmetric filtering to focus on genuinely uncertain samples. Classification samples are retained when they achieve 1--7 correct rollouts (partial success), while discovery samples are retained when they achieve 1--3 correct rollouts (harder threshold reflecting the higher difficulty of discovery prediction). Samples that always succeed or always fail provide no learning signal and are discarded. We maintain a roughly 9:1 ratio between classification and discovery samples to balance the training distribution.

We employ Group Relative Policy Optimization~\cite{shao2024deepseekmath} for training. For each query, $G$ rollouts form a group $\mathcal{G} = \{s^{(1)}, \ldots, s^{(G)}\}$. We compute group-relative advantages normalized by the standard deviation:
\begin{equation}
\hat{A}^{(i)} = \frac{R(s^{(i)}) - \mathrm{mean}\!\left(\{R(s^{(j)})\}_{j=1}^{G}\right)}{\mathrm{std}\!\left(\{R(s^{(j)})\}_{j=1}^{G}\right)}
\end{equation}
The reward function $R(s)$ assigns $+1$ for correct classification or discovery decisions, and $+0.1$ for maintaining proper output structure (valid \texttt{<think>...</think><answer>...</answer>} format), with 0 otherwise. The clipped surrogate objective $\mathcal{J}_{\text{GRPO}}$ is as follows:
\begin{equation}
\mathbb{E}_{\mathcal{G}} \!\left[ \frac{1}{G}\sum_{i=1}^{G} \frac{1}{T_i}\sum_{t=1}^{T_i} \min\!\left( \rho_t^{(i)} \hat{A}^{(i)},\, \mathrm{clip}\!\left(\rho_t^{(i)}, 1{-}\epsilon, 1{+}\epsilon\right) \hat{A}^{(i)} \right) \right]
\end{equation}
where $\rho_t^{(i)} = \pi_\theta(s_t^{(i)} \mid s_{<t}^{(i)}, \mathbf{x}_q, \mathcal{R}) \,/\, \pi_{\theta_{\text{old}}}(s_t^{(i)} \mid s_{<t}^{(i)}, \mathbf{x}_q, \mathcal{R})$ is the importance sampling ratio and $\epsilon$ is the clipping range. 

Through two-stage training, DeepTaxon learns to convert high-recall retrieval into high-precision decisions, systematically bridging the gap between retrieval potential and actual performance for unified classification and discovery.

\begin{table}[t]
\centering\small
\caption{Evaluation datasets. Classification uses domain-specific indices. Discovery tests against the iNaturalist index using novel (non-overlapping) categories.}
\label{tab:eval_data}
\begin{tabular*}{\columnwidth}{@{\extracolsep{\fill}}lcccc@{}}
\toprule
 & \multicolumn{2}{c}{Classification} & \multicolumn{2}{c}{Discovery} \\
\cmidrule(lr){2-3} \cmidrule(lr){4-5}
Dataset & Classes & Samples & Novel & Samples \\
\midrule
iNaturalist~\cite{vanhorn2018inaturalist} & 10,000 & 10,000 & --- & --- \\
Flowers102~\cite{nilsback2008automated} & 102 & 6,149 & 60 & 2,477 \\
Butterfly-200~\cite{butterfly200} & 200 & 15,009 & 130 & 1,000 \\
Stanford Dogs~\cite{khosla2011novel} & 120 & 8,580 & 3 & 175 \\
Food-101~\cite{bossard2014food} & 101 & 25,250 & 101 & 1,000 \\
Stanford Cars~\cite{krause20133d} & 196 & 8,041 & 194 & 1,000 \\
FGVC-Aircraft~\cite{maji2013fine} & 196 & 3,333 & 196 & 3,333 \\
\bottomrule
\end{tabular*}
\end{table}

\section{Experiments}
\label{sec:experimental}

\begin{table*}[t]
\centering\small
\caption{Main results on iNaturalist classification and out-of-distribution discovery (RQ1). Classification reports Accuracy (\%). Discovery reports the rate (\%) of correctly identifying samples as novel. OOD methods are shown at two threshold settings: Cls-Preserving (preserves classification) and Disc-Optimized$^\dagger$ ($\leq$5\% relative drop).}
\label{tab:main_results}
\begin{tabular*}{\textwidth}{@{\extracolsep{\fill}}lllccccccc@{}}
\toprule
 & & & \multicolumn{1}{c}{Classification} & \multicolumn{6}{c}{Discovery} \\
\cmidrule(lr){4-4} \cmidrule(lr){5-10}
Category & Model & Method & iNaturalist & Flowers & Butterfly & Dogs & Food & Cars & Aircraft \\
\midrule
\multirow{8}{*}{\shortstack{OOD\\(threshold)}}
& \multirow{4}{*}{ResNet-50} & MSP (Cls-Preserving) & 19.92 & 0.0 & 0.0 & 0.0 & 0.0 & 0.0 & 0.0 \\
& & MSP (Disc-Optimized$^\dagger$) & 19.07 & 11.2 & 13.6 & 30.3 & 33.4 & 20.8 & 43.5 \\
& & VIM (Cls-Preserving) & 19.92 & 0.0 & 0.0 & 0.0 & 0.0 & 0.0 & 0.0 \\
& & VIM (Disc-Optimized$^\dagger$) & 18.93 & 8.6 & 18.9 & 6.9 & 18.3 & 0.6 & 3.6 \\
\cmidrule(lr){2-10}
& \multirow{4}{*}{ViT-H} & MSP (Cls-Preserving) & 39.30 & 1.7 & 0.4 & 1.7 & 23.2 & 46.4 & 59.3 \\
& & MSP (Disc-Optimized$^\dagger$) & 38.24 & 26.2 & 9.2 & 33.1 & 74.5 & 77.0 & 92.9 \\
& & VIM (Cls-Preserving) & 39.30 & 0.0 & 0.0 & 0.0 & 0.1 & 9.9 & 8.1 \\
& & VIM (Disc-Optimized$^\dagger$) & 38.09 & 20.3 & 7.7 & 32.0 & 68.7 & 98.1 & 98.8 \\
\midrule
\multirow{3}{*}{DeepTaxon}
& \multirow{3}{*}{Qwen2.5-VL-7B} & Zero-Shot$^\ddagger$ & 24.87 & 16.3 & 13.2 & 33.7 & 95.7 & 99.7 & 97.3 \\
& & +SFT & 57.80 & 44.6 & 12.5 & 34.9 & \textbf{95.9} & 99.7 & \textbf{99.9} \\
& & +GRPO & \textbf{60.07} & \textbf{49.4} & \textbf{15.6} & \textbf{42.9} & 95.7 & \textbf{99.8} & \textbf{99.9} \\
\bottomrule
\multicolumn{10}{l}{\footnotesize $^\dagger$Disc-Optimized: threshold tuned on validation set with $\leq$5\% relative classification drop.} \\
\multicolumn{10}{l}{\footnotesize $^\ddagger$Zero-Shot: Qwen2.5-VL-7B with retrieval augmentation but without fine-tuning (same prompt as trained models).} \\
\end{tabular*}
\end{table*}

We evaluate DeepTaxon on large-scale species identification and open-set discovery. Our experiments address five research questions:
\begin{itemize}[nosep,leftmargin=*]
    \item \textbf{RQ1:} How effective is DeepTaxon for unified classification and discovery compared to OOD detection methods?
    \item \textbf{RQ2:} How does test-time scaling with retrieval parameters $(k, n)$ affect performance?
    \item \textbf{RQ3:} Can the framework generalize to new domains by simply changing the retrieval index?
    \item \textbf{RQ4:} How do different retrieval encoders affect the retrieval-decision gap?
    \item \textbf{RQ5:} How does DeepTaxon perform on cross-domain classification and discovery?
\end{itemize}

\subsection{Experimental Setup}
\label{subsec:setup}

\subsubsection{Datasets}

\textbf{Training Data.} We use a subset of iNaturalist~\cite{vanhorn2018inaturalist} containing 500,000 images across 10,000 species. We reserve 100,000 images for synthetic data generation through the pipeline described in Section~\ref{subsec:synthetic_data}, yielding 76,621 SFT samples (59,709 classification, 16,912 discovery). The remaining 400,000 images serve as the retrieval index.

For reinforcement learning (RL), we construct four training settings to enhance cross-domain generalization, using ImageNet-21k subset~\cite{ridnik2021imagenet21k,chittyvenkata2025imagenetthink} as an additional data source:
\begin{itemize}[nosep,leftmargin=*]
\item \textbf{Setting 1}: iNaturalist queries $\rightarrow$ iNaturalist index (in-domain hard samples)
\item \textbf{Setting 2}: iNaturalist queries $\rightarrow$ ImageNet-21k index (OOD discovery)
\item \textbf{Setting 3}: ImageNet-21k queries $\rightarrow$ ImageNet-21k index (cross-domain classification)
\item \textbf{Setting 4}: ImageNet-21k queries $\rightarrow$ iNaturalist index (cross-domain discovery)
\end{itemize}
After filtering with the procedure in Section~\ref{subsec:rl}, Settings 1, 3 yield approximately 10,000 hard samples. We further augment with 500 cross-domain discovery samples (250 from Setting 2 and 250 from Setting 4) to strengthen cross-domain discovery capability. In total, we construct 10,500 hard RL training samples (8,994 classification, 1,506 discovery).

\textbf{Evaluation Data.} We evaluate on two complementary tasks: classification and discovery (Table~\ref{tab:eval_data}). For classification, the primary evaluation uses the iNaturalist evaluation set (10,000 species, one held-out image per species). Critically, evaluation images are drawn from the iNaturalist mini eval split (50k images), from which we sample 10k (one per species). These evaluation images are strictly disjoint from both the 400k retrieval index and the 100k images used for synthetic data generation, ensuring no overlap between training queries and evaluation queries. Cross-domain classification uses six additional fine-grained datasets with each dataset's training split as retrieval index. To evaluate discovery capability, we test whether the model correctly identifies samples whose categories are absent from the iNaturalist retrieval index. We use an MLLM to automatically determine whether each dataset's category labels overlap with iNaturalist species, followed by manual verification, and exclude overlapping categories from discovery evaluation. Flowers102, Butterfly-200, and Stanford Dogs are identified as partially overlapping, so we retain only non-overlapping categories (Novel). Food-101, Stanford Cars, and FGVC-Aircraft show no overlap and are used entirely. For efficiency, some datasets are evaluated on a randomly sampled subset for discovery, so not all novel categories may be covered in the sampled images.

\subsubsection{Evaluation Metrics}

The model outputs \texttt{[Classification]: <label>} or \texttt{[Discovery]} for each query. For classification, we report accuracy as the proportion of queries where the predicted label matches the ground truth. For discovery, we report the discovery rate as the proportion of novel-label queries where the model correctly outputs \texttt{[Discovery]} instead of selecting a candidate. The complete prompt template and qualitative examples are provided in Appendices~\ref{app:prompt} and~\ref{app:casestudy}.

\subsubsection{Implementation Details}

We use OpenCLIP~\cite{ilharco2021openclip} vision encoders (ViT-B/32, ViT-L/14, ViT-H/14, all pretrained on LAION-2B~\cite{schuhmann2022laion5b}; abbreviated ViT-B, ViT-L, ViT-H respectively) for retrieval with FAISS indexing~\cite{johnson2019billion}. The reasoning model is Qwen2.5-VL-7B-Instruct~\cite{bai2025qwen25vl} fine-tuned with LoRA~\cite{hu2021lora} at rank 128. Both stages use the Adam optimizer~\cite{kingma2015adam}. For supervised fine-tuning, we use learning rate $10^{-4}$ with cosine schedule and train for one epoch. For GRPO, we reduce learning rate to $10^{-5}$ and train for two epochs with $G=8$ rollouts per query. Training uses 8 H20 GPUs with DeepSpeed ZeRO-3~\cite{rajbhandari2020zero}. Unless otherwise specified, we report results with ViT-H encoder at $k=16$ and $n=4$.

\begin{table*}[t]
\centering\small
\caption{Test-time scaling on iNaturalist classification (RQ2). Accuracy (\%). Rows vary candidate count $k$, and columns vary exemplars per species $n$. Bold indicates the best result.}
\label{tab:scaling}
\begin{tabular*}{\textwidth}{@{\extracolsep{\fill}}lccccccccc@{}}
\toprule
 & \multicolumn{3}{c}{Zero-Shot} & \multicolumn{3}{c}{SFT} & \multicolumn{3}{c}{DeepTaxon} \\
\cmidrule(lr){2-4} \cmidrule(lr){5-7} \cmidrule(lr){8-10}
$k$ & $n{=}1$ & $n{=}2$ & $n{=}4$ & $n{=}1$ & $n{=}2$ & $n{=}4$ & $n{=}1$ & $n{=}2$ & $n{=}4$ \\
\midrule
4  & 32.75 & 31.63 & 28.63 & 43.72 & 46.84 & 47.85 & 44.45 & 47.86 & 49.57 \\
8  & 30.58 & 27.97 & 28.33 & 47.12 & 51.19 & 53.17 & 49.45 & 53.62 & 55.66 \\
12 & 28.31 & 27.82 & 27.55 & 49.59 & 54.67 & 56.74 & 51.36 & 56.09 & 58.79 \\
16 & 28.25 & 26.33 & 24.87 & 50.81 & 55.65 & 57.80 & 52.70 & 57.57 & 60.07 \\
24 & 30.52 & 31.60 & 21.48 & 51.98 & 56.81 & 59.10 & 53.43 & 58.58 & 61.39 \\
32 & 31.65 & 30.89 & 21.62 & 51.70 & 57.12 & 59.33 & 53.16 & 59.17 & \textbf{61.63} \\
\bottomrule
\end{tabular*}
\end{table*}

\begin{table}[t]
\centering
\caption{Cross-domain classification accuracy (RQ3). Accuracy (\%). ``w/o Retrieval'' methods rely solely on parametric knowledge, while ``w/ Retrieval'' methods use domain-specific retrieval indices. Bold indicates best per dataset.}
\label{tab:crossdomain}
\resizebox{\columnwidth}{!}{%
\begin{tabular}{llcccccc}
\toprule
Category & Method & Flowers & Butterfly & Dogs & Food & Cars & Aircraft \\
\midrule
\multirow{2}{*}{w/o Retrieval} & Qwen2.5-VL-32B & 36.02 & 18.78 & 5.89 & 27.89 & 1.95 & 2.22 \\
& Qwen2.5-VL-72B & 48.43 & 20.11 & 17.36 & 49.55 & 3.53 & 4.23 \\
\midrule
\multirow{2}{*}{w/ Retrieval} & Qwen2.5-VL-7B & 40.59 & 59.31 & 51.23 & 28.90 & 83.60 & 27.63 \\
& DeepTaxon & \textbf{99.17} & \textbf{80.35} & \textbf{84.10} & \textbf{89.98} & \textbf{93.02} & \textbf{67.38} \\
\bottomrule
\end{tabular}%
}
\end{table}

\subsection{Main Results (RQ1)}
\label{subsec:main_results}

Table~\ref{tab:main_results} compares DeepTaxon with two post-hoc OOD scoring methods, MSP~\cite{hendrycks2017baseline} and VIM~\cite{wang2022vim}, under two threshold settings: \textit{Cls-Preserving} (preserves full closed-set classification accuracy) and \textit{Disc-Optimized} (maximizes discovery with $\leq$5\% relative classification drop). We additionally include ResNet-50~\cite{he2016deep} as a weaker encoder baseline. Detailed baseline descriptions, threshold definitions, and additional results with ViT-B and ViT-L encoders are provided in Appendix~\ref{app:ood_full}.

According to Table~\ref{tab:main_results}, we have the following observations.
(1) Traditional OOD methods exhibit a fundamental classification-discovery trade-off. At the Cls-Preserving threshold, discovery collapses to near zero across most domains (e.g., 0.0\% on Flowers and Butterfly for both MSP and VIM with ViT-H). The Disc-Optimized threshold recovers discovery on semantically distant domains (ViT-H MSP: 92.9\% on Aircraft, 77.0\% on Cars), but at the cost of classification accuracy, and discovery on biologically-related domains remains limited (Dogs 33.1\%, Flowers 26.2\%, Butterfly 9.2\%).
(2) DeepTaxon unifies classification and discovery without this trade-off. SFT dramatically improves classification from 24.87\% (zero-shot) to 57.80\% while enhancing discovery on biologically-related domains, with Flowers increasing from 16.3\% to 44.6\% and Dogs from 33.7\% to 34.9\%. On semantically distant domains, discovery is consistently near-perfect (Food $\geq$95.7\%, Cars $\geq$99.7\%, Aircraft $\geq$97.3\%).
(3) GRPO further improves both tasks. Classification increases from 57.80\% to 60.07\%, narrowing the retrieval-decision gap toward the Pass@$k$ ceiling. Discovery on hard biological domains also benefits, with Flowers gaining 4.8 points (44.6\%$\rightarrow$49.4\%) and Dogs gaining 8.0 points (34.9\%$\rightarrow$42.9\%), demonstrating that reinforcement learning on challenging samples is essential for fine-grained discrimination.

\subsection{Test-Time Scaling (RQ2)}
\label{subsec:scaling}

A distinctive capability of DeepTaxon is test-time scaling. The ability to trade additional inference computation for improved accuracy by adjusting the retrieval parameters $k$ (number of candidate species) and $n$ (exemplars per species). We evaluate this capability by varying $(k, n)$ configurations and comparing three model variants: zero-shot (Qwen2.5-VL-7B with retrieval but no fine-tuning), SFT (after supervised fine-tuning), and DeepTaxon (after GRPO training).

According to Table~\ref{tab:scaling}, we observe: (1) Zero-shot performance generally degrades as $k$ increases, particularly with larger $n$, dropping from 32.75\% at $(k{=}4, n{=}1)$ to 21.48\% at $(k{=}24, n{=}4)$ and 21.62\% at $(k{=}32, n{=}4)$. Without task-specific training, additional candidates introduce noise that confuses the model rather than providing useful comparative information. (2) Both SFT and DeepTaxon exhibit consistent scaling improvements with larger $(k, n)$ values across the full range. DeepTaxon improves from 44.45\% at $(k{=}4, n{=}1)$ to 61.63\% at $(k{=}32, n{=}4)$, a gain of 17.18 points purely from expanding the reasoning environment. Notably, performance continues improving beyond $k{=}16$: from 60.07\% at $(k{=}16, n{=}4)$ to 61.63\% at $(k{=}32, n{=}4)$, demonstrating that our training approach successfully teaches the model to leverage richer comparative information even at larger scales. (3) GRPO consistently outperforms SFT across all $(k, n)$ configurations, with the gap widening at larger scales (2.30 points at $k{=}32, n{=}4$), demonstrating that reinforcement learning on hard samples improves reasoning quality. This test-time scaling property enables flexible deployment, as users can adjust $(k, n)$ based on accuracy requirements and computational constraints.

\subsection{Cross-Domain Generalization (RQ3)}
\label{subsec:crossdomain}

A key advantage of retrieval-augmented approaches is the ability to generalize to new domains by simply changing the retrieval index without retraining. We evaluate this capability by using each dataset's training split as the retrieval index, comparing DeepTaxon against both retrieval-free MLLMs (Qwen2.5-VL-32B/72B) and the zero-shot retrieval baseline (Qwen2.5-VL-7B with retrieval but without fine-tuning).

According to Table~\ref{tab:crossdomain}, we have the following observations.
(1) Retrieval-free MLLMs fail on fine-grained domains even with 10$\times$ larger parameters. Qwen2.5-VL-72B achieves only 4.23\% on Aircraft and 3.53\% on Cars, demonstrating that implicit parametric knowledge alone cannot support fine-grained recognition at scale.
(2) Retrieval augmentation enables dramatic improvements. Even a 7B model with retrieval reaches 83.60\% on Cars, far exceeding the 72B model without retrieval (3.53\%), confirming that external visual evidence is essential for fine-grained discrimination.
(3) DeepTaxon's comparative reasoning training further amplifies these gains, improving over Qwen2.5-VL-7B (without fine-tuning) by 58.58 points on Flowers (99.17\% vs 40.59\%) and 39.75 points on Aircraft (67.38\% vs 27.63\%), demonstrating that our training approach effectively teaches the model to leverage retrieved evidence.

\begin{table*}[t]
\centering\small
\caption{Impact of retrieval encoder on classification accuracy (RQ4). Accuracy (\%) across all datasets. Top-1 selects the most similar species directly. DeepTaxon applies comparative reasoning over $k=16$ candidates with $n=4$ exemplars.}
\label{tab:retriever}
\begin{tabular*}{\textwidth}{@{\extracolsep{\fill}}llccccccc@{}}
\toprule
Encoder & Method & iNaturalist & Flowers & Butterfly & Dogs & Food & Cars & Aircraft \\
\midrule
\multirow{2}{*}{ViT-B} & Top-1 & 19.51 & 88.43 & 53.84 & 57.61 & 79.47 & 81.33 & 41.31 \\
& DeepTaxon & 47.60 & 98.78 & 78.68 & 82.26 & 89.65 & 91.33 & 63.00 \\
\midrule
\multirow{2}{*}{ViT-L} & Top-1 & 32.04 & 96.51 & 63.70 & 69.63 & 89.68 & 90.57 & 55.68 \\
& DeepTaxon & 58.87 & 99.08 & \textbf{80.63} & 83.77 & 89.88 & \textbf{93.28} & 66.09 \\
\midrule
\multirow{2}{*}{ViT-H} & Top-1 & 34.64 & 97.20 & 64.94 & 72.16 & \textbf{90.98} & 89.80 & 58.56 \\
& DeepTaxon & \textbf{60.07} & \textbf{99.17} & 80.35 & \textbf{84.10} & 89.98 & 93.02 & \textbf{67.38} \\
\bottomrule
\end{tabular*}
\end{table*}

\subsection{Retriever Analysis (RQ4)}
\label{subsec:retriever}

DeepTaxon is designed to be retriever-agnostic. Our reasoning model should improve upon any retrieval backbone by bridging the retrieval-decision gap. We validate this by comparing DeepTaxon against the Top-1 retrieval baseline (selecting the most similar species directly) across three encoder scales (ViT-B, ViT-L, ViT-H).

According to Table~\ref{tab:retriever}, we have the following observations.
(1) DeepTaxon consistently outperforms Top-1 retrieval across all three encoder scales, confirming that it is retriever-agnostic. DeepTaxon improves in 20 out of 21 encoder-dataset combinations, with gains up to +28.09 points. Notably, even with the weakest encoder (ViT-B), DeepTaxon achieves 47.60\% on iNaturalist, surpassing ViT-H Top-1 (34.64\%), demonstrating that comparative reasoning can partially compensate for weaker retrieval quality.
(2) Performance scales with encoder strength. As the retrieval encoder scales from ViT-B to ViT-L to ViT-H, DeepTaxon performance generally improves. On iNaturalist, accuracy increases from 47.60\% to 58.87\% to 60.07\%.
(3) The improvements are most pronounced on the four biological datasets. On iNaturalist, DeepTaxon improves over Top-1 by 25.43--28.09 points across all encoders. On Butterfly-200 and Stanford Dogs, gains range from 11.94 to 24.84 points, as these fine-grained biological domains require detailed comparative reasoning that nearest-neighbor matching cannot provide. On Flowers102, where Top-1 already achieves above 88\%, DeepTaxon pushes accuracy to near-perfect levels (98.78--99.17\%).

\begin{figure}[t]
\centering
\includegraphics[width=\columnwidth]{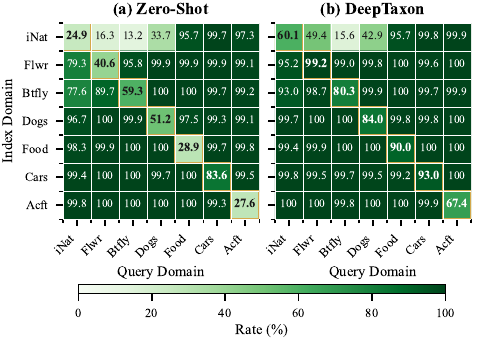}
\caption{Cross-domain evaluation matrices (RQ5). Rows represent retrieval index domains, columns represent query domains. Diagonal (boxed) = classification accuracy, off-diagonal = discovery rate.}
\label{fig:confusion_heatmap}
\end{figure}

\subsection{Unified Classification and Discovery (RQ5)}
\label{subsec:unified}

We evaluate DeepTaxon's unified capability through cross-domain evaluation matrices (Figure~\ref{fig:confusion_heatmap}), where rows represent retrieval index domains and columns represent query domains. Diagonal entries measure classification accuracy, and off-diagonal entries measure discovery rates (query categories are always disjoint from the retrieval index categories). We randomly sample up to 1,000 queries from each dataset for efficiency. Detailed numeric values are provided in Appendix~\ref{app:confusion_tables}.

According to Figure~\ref{fig:confusion_heatmap}, we have the following observations.
(1) Cross-domain discovery improves consistently. DeepTaxon achieves higher off-diagonal discovery rates than the zero-shot retrieval baseline on nearly all domain pairs, demonstrating that synthetic data construction, SFT, and RL are effective for the discovery task. Discovery rates are near-perfect when domains are semantically distinct (e.g., FGVC-Aircraft queries against any biological index achieve $\geq$99.8\%), and also improve on challenging biologically-related pairs, with Flowers queries against the iNaturalist index increasing from 16.3\% to 49.4\% and Dogs queries from 33.7\% to 42.9\%.
(2) Open-set classification generalizes across domains. Despite being trained exclusively on iNaturalist in-domain data, DeepTaxon significantly improves diagonal classification accuracy across all domains by simply swapping the retrieval index. Flowers102 improves from 40.6\% to 99.2\% (+58.6), FGVC-Aircraft from 27.6\% to 67.4\% (+39.8), iNaturalist from 24.9\% to 60.1\% (+35.2), and Butterfly-200 from 59.3\% to 80.3\% (+21.0).
(3) Biological query domains benefit most. Examining the first four columns (iNaturalist, Flowers, Butterfly, Dogs as queries), DeepTaxon improves both classification and discovery across all index domains. This is particularly notable because fine-grained biological queries are the most challenging, requiring detailed comparative reasoning to distinguish visually similar species or to recognize novel ones.

\section{Conclusion}
\label{sec:conclusion}

We present DeepTaxon, a retrieval-augmented multimodal reasoning framework that unifies species identification and discovery through comparative reasoning over retrieved visual evidence. By redefining discovery as an explicit, retrieval-based decision problem, each retrieval naturally provides supervision for both tasks without manual annotation, and every prediction is grounded in interpretable visual comparisons. Experiments on a 10K-class benchmark and six cross-domain datasets confirm consistent improvements in identification, discovery, test-time scaling, and cross-domain generalization. Beyond biodiversity, we believe this paradigm can extend to other fine-grained recognition domains that require both open-set generalization and interpretable decision-making.

\section{Limitations and Ethical Considerations}
\label{sec:limitations_ethics}

DeepTaxon has several limitations. Inference latency scales with $(k, n)$, creating an accuracy-efficiency trade-off. Classification accuracy is bounded by the Pass@$k$ retrieval ceiling. Discovery in biodiversity domains remains challenging due to high visual similarity. Regarding ethical considerations, the system's decisions are constrained by the coverage and quality of the retrieval index, so careful index curation is important for reliable deployment. The interpretable reasoning chains allow domain experts to verify the visual evidence behind each decision. We emphasize that automated identification should assist rather than replace taxonomic expertise.

\bibliographystyle{ACM-Reference-Format}
\bibliography{reference}

\appendix

\section{Pass@$k$ and DeepTaxon Accuracy}
\label{app:passk_data}

Table~\ref{tab:passk_raw} reports the numerical values behind the Pass@$k$ curves in Figure~\ref{fig:teaserfigure}. Pass@$k$ denotes the retrieval ceiling (probability that the ground-truth species appears among the top-$k$ retrieved distinct species), and DeepTaxon reports the classification accuracy after comparative reasoning. Results are shown for three OpenCLIP encoders (ViT-B, ViT-L, ViT-H) on iNaturalist-10K. Across all encoders, DeepTaxon accuracy increases monotonically with $k$, confirming effective test-time scaling through comparative reasoning over progressively richer candidate sets.

\begin{table*}[htbp]
\centering\small
\caption{Pass@$k$ (retrieval ceiling) and DeepTaxon classification accuracy (\%) on iNaturalist-10K across three retrieval encoders. DeepTaxon uses $n{=}4$ exemplars per candidate species.}
\label{tab:passk_raw}
\begin{tabular*}{\textwidth}{@{\extracolsep{\fill}}ccccccc@{}}
\toprule
 & \multicolumn{2}{c}{ViT-B} & \multicolumn{2}{c}{ViT-L} & \multicolumn{2}{c}{ViT-H} \\
\cmidrule(lr){2-3} \cmidrule(lr){4-5} \cmidrule(lr){6-7}
$k$ & Pass@$k$ & DeepTaxon & Pass@$k$ & DeepTaxon & Pass@$k$ & DeepTaxon \\
\midrule
1  & 19.51 & 19.29 & 32.04 & 31.73 & 34.64 & 34.26 \\
2  & 27.84 & 20.90 & 44.50 & 40.61 & 46.87 & 42.61 \\
3  & 33.82 & 30.83 & 51.77 & 44.93 & 54.34 & 46.88 \\
4  & 38.04 & 33.67 & 56.66 & 47.90 & 59.35 & 49.57 \\
5  & 41.60 & 36.22 & 59.83 & 49.46 & 63.22 & 51.50 \\
6  & 44.51 & 38.11 & 62.75 & 51.25 & 65.98 & 53.10 \\
7  & 46.95 & 39.70 & 65.32 & 52.70 & 68.38 & 54.69 \\
8  & 48.96 & 41.32 & 67.44 & 53.84 & 70.17 & 55.66 \\
9  & 50.63 & 42.36 & 69.04 & 55.05 & 71.97 & 56.59 \\
10 & 52.12 & 43.39 & 70.36 & 55.81 & 73.49 & 57.66 \\
11 & 53.74 & 44.67 & 71.67 & 56.50 & 74.78 & 58.14 \\
12 & 55.04 & 45.47 & 72.84 & 57.40 & 76.05 & 58.79 \\
13 & 56.19 & 46.03 & 73.81 & 57.64 & 77.22 & 59.25 \\
14 & 57.41 & 46.60 & 74.83 & 58.10 & 78.16 & 59.59 \\
15 & 58.63 & 47.02 & 75.67 & 58.58 & 78.91 & 59.74 \\
16 & 59.63 & 47.60 & 76.60 & 58.87 & 79.81 & 60.07 \\
24 & 65.64 & 50.52 & 81.20 & 60.23 & 83.84 & 61.39 \\
32 & 69.94 & 51.70 & 84.24 & 60.63 & 86.52 & 61.63 \\
\bottomrule
\end{tabular*}
\end{table*}

\section{Prompt Template}
\label{app:prompt}

We use the following prompt template for both synthetic data generation and inference. The prompt instructs the model to perform comparative analysis between the query image and retrieved reference images.

{\footnotesize
\begin{verbatim}
# Role
You are a professional classification and discovery expert.

# Task
You are given:
* 1 query image (unknown object).
* N reference entries, each consisting of one or more
  images and a full taxonomy.
Your goal is to identify whether the query belongs to one
of the reference objects or represents a new object.
Analyze the visual characteristics of the query image by
comparing it with the reference images, then decide ONE
of the following:
* [Classification] - the query belongs to one of the
  reference objects. Output the taxonomy of that reference.
* [Discovery] - treat it as a new object.

# Output Format
[Classification]: taxonomy
or
[Discovery]

# Input
Query image: <image>
Reference 1:
R1I1: <image> R1I2: <image> ...
taxonomy: ...
Reference 2:
R2I1: <image> R2I2: <image> ...
taxonomy: ...
\end{verbatim}
}

The model generates a reasoning chain analyzing visual similarities and differences before outputting the final decision.

\section{OOD Detection Baseline Details}
\label{app:ood_full}

We compare against two representative post-hoc OOD scoring methods that require no retraining of the classification backbone.

\textbf{MSP}~\cite{hendrycks2017baseline} uses the maximum softmax probability as an in-distribution confidence score. A sample is classified normally if its confidence exceeds the threshold, and flagged as OOD (discovery) otherwise.

\textbf{VIM}~\cite{wang2022vim} combines logit-space and feature-space information by computing a virtual logit from the residual of the feature vector projected onto the principal subspace of training features, providing a more expressive OOD score than softmax-based methods.

\textbf{Training Data.} To ensure a fair comparison, the OOD classification backbones are trained on the same query images used to construct DeepTaxon's SFT and RL training data. Note that OOD backbones can only leverage these images for classification training, whereas DeepTaxon jointly learns both classification and discovery from the same data.

\textbf{Threshold Settings.} We evaluate each method under two threshold settings to reveal both the structural limitation and the practical trade-off of threshold-based OOD detection. The \textit{Cls-Preserving} threshold uses the minimum confidence score observed in the training set as the rejection boundary, preserving the full closed-set classification accuracy and exposing each method's inherent discovery baseline. The \textit{Disc-Optimized} threshold is selected on an independent validation set to find the strictest threshold whose classification accuracy does not drop below 95\% of the Cls-Preserving baseline, i.e., at most 5\% relative degradation. The validation set is constructed by per-class uniform sampling from the original iNaturalist and ImageNet-21k training sets, fully disjoint from our training and test data.

Table~\ref{tab:ood_vitb_vitl} provides additional OOD detection results with ViT-L and ViT-B encoders, complementing the ViT-H and ResNet-50 results in Table~\ref{tab:main_results}.

\begin{table*}[htbp]
\centering\small
\caption{OOD detection results with ViT-L and ViT-B encoders. Same format as Table~\ref{tab:main_results}.}
\label{tab:ood_vitb_vitl}
\begin{tabular*}{\textwidth}{@{\extracolsep{\fill}}llccccccc@{}}
\toprule
Model & Method & iNaturalist & Flowers & Butterfly & Dogs & Food & Cars & Aircraft \\
\midrule
\multirow{4}{*}{ViT-L} & MSP (Cls-Preserving) & 45.86 & 0.0 & 0.0 & 0.0 & 2.3 & 40.3 & 21.5 \\
& MSP (Disc-Optimized$^\dagger$) & 44.74 & 19.7 & 6.8 & 32.6 & 76.8 & 96.0 & 94.0 \\
& VIM (Cls-Preserving) & 45.86 & 0.0 & 0.0 & 0.0 & 0.0 & 0.1 & 0.0 \\
& VIM (Disc-Optimized$^\dagger$) & 45.86 & 0.2 & 0.4 & 0.0 & 48.3 & 99.0 & 95.4 \\
\cmidrule(lr){1-9}
\multirow{4}{*}{ViT-B} & MSP (Cls-Preserving) & 11.05 & 0.0 & 0.0 & 0.0 & 0.0 & 1.5 & 0.0 \\
& MSP (Disc-Optimized$^\dagger$) & 10.75 & 1.5 & 0.6 & 41.7 & 52.4 & 99.8 & 88.5 \\
& VIM (Cls-Preserving) & 11.05 & 0.0 & 0.0 & 0.0 & 0.0 & 0.0 & 0.0 \\
& VIM (Disc-Optimized$^\dagger$) & 10.71 & 4.5 & 4.1 & 6.9 & 64.7 & 99.9 & 99.0 \\
\bottomrule
\end{tabular*}

\noindent\footnotesize $^\dagger$Disc-Optimized: threshold tuned on validation set with $\leq$5\% relative classification drop.
\end{table*}

\section{Cross-Domain Evaluation Matrices}
\label{app:confusion_tables}

Tables~\ref{tab:confusion_zeroshot_app} and~\ref{tab:confusion_deeptaxon_app} provide the complete numerical values for the cross-domain evaluation matrices visualized in Figure~\ref{fig:confusion_heatmap}. Diagonal entries measure classification accuracy within each domain, while off-diagonal entries measure the discovery rate when query categories are always disjoint from the retrieval index categories. We randomly sample up to 1,000 queries from each dataset for efficiency. Comparing the two tables, DeepTaxon improves diagonal classification accuracy across all seven domains and raises off-diagonal discovery rates on the majority of cross-domain pairs, demonstrating that the unified retrieval-augmented reasoning framework simultaneously strengthens both tasks.

\begin{table}[t]
\centering
\caption{Cross-domain evaluation matrix for Zero-Shot baseline. Diagonal: classification accuracy (\%). Off-diagonal: discovery rate (\%).}
\label{tab:confusion_zeroshot_app}
\resizebox{\columnwidth}{!}{
\begin{tabular}{lccccccc}
\toprule
Index $\downarrow$ Query $\rightarrow$ & iNaturalist & Flowers & Butterfly & Dogs & Food & Cars & Aircraft \\
\midrule
iNaturalist & \textbf{24.87} & 16.26 & 13.20 & 33.71 & 95.70 & 99.67 & 97.32 \\
Flowers102 & 79.30 & \textbf{40.59} & 95.80 & 99.90 & 99.90 & 99.90 & 99.10 \\
Butterfly-200 & 77.60 & 89.70 & \textbf{59.31} & 100 & 100 & 99.70 & 99.20 \\
Stanford Dogs & 96.70 & 100 & 99.90 & \textbf{51.23} & 97.50 & 99.30 & 99.10 \\
Food-101 & 98.30 & 99.90 & 100 & 100 & \textbf{28.90} & 99.70 & 99.80 \\
Stanford Cars & 99.40 & 100 & 100 & 99.70 & 100 & \textbf{83.60} & 99.50 \\
FGVC-Aircraft & 99.80 & 100 & 100 & 100 & 100 & 99.30 & \textbf{27.63} \\
\bottomrule
\end{tabular}
}
\end{table}

\begin{table}[t]
\centering
\caption{Cross-domain evaluation matrix for DeepTaxon. Diagonal: classification accuracy (\%). Off-diagonal: discovery rate (\%).}
\label{tab:confusion_deeptaxon_app}
\resizebox{\columnwidth}{!}{
\begin{tabular}{lccccccc}
\toprule
Index $\downarrow$ Query $\rightarrow$ & iNaturalist & Flowers & Butterfly & Dogs & Food & Cars & Aircraft \\
\midrule
iNaturalist & \textbf{60.07} & 49.37 & 15.60 & 42.85 & 95.73 & 99.82 & 99.90 \\
Flowers102 & 95.20 & \textbf{99.23} & 99.00 & 99.80 & 100 & 99.60 & 100 \\
Butterfly-200 & 93.00 & 98.70 & \textbf{80.35} & 99.90 & 100 & 99.70 & 99.90 \\
Stanford Dogs & 99.70 & 100 & 100 & \textbf{83.98} & 99.80 & 99.80 & 100 \\
Food-101 & 99.40 & 99.90 & 100 & 100 & \textbf{89.98} & 100 & 100 \\
Stanford Cars & 99.80 & 99.50 & 99.70 & 99.50 & 99.20 & \textbf{93.02} & 100 \\
FGVC-Aircraft & 100 & 100 & 99.80 & 100 & 100 & 99.90 & \textbf{67.38} \\
\bottomrule
\end{tabular}
}
\end{table}

\begin{figure*}[t]
    \centering
    \includegraphics[width=0.95\textwidth]{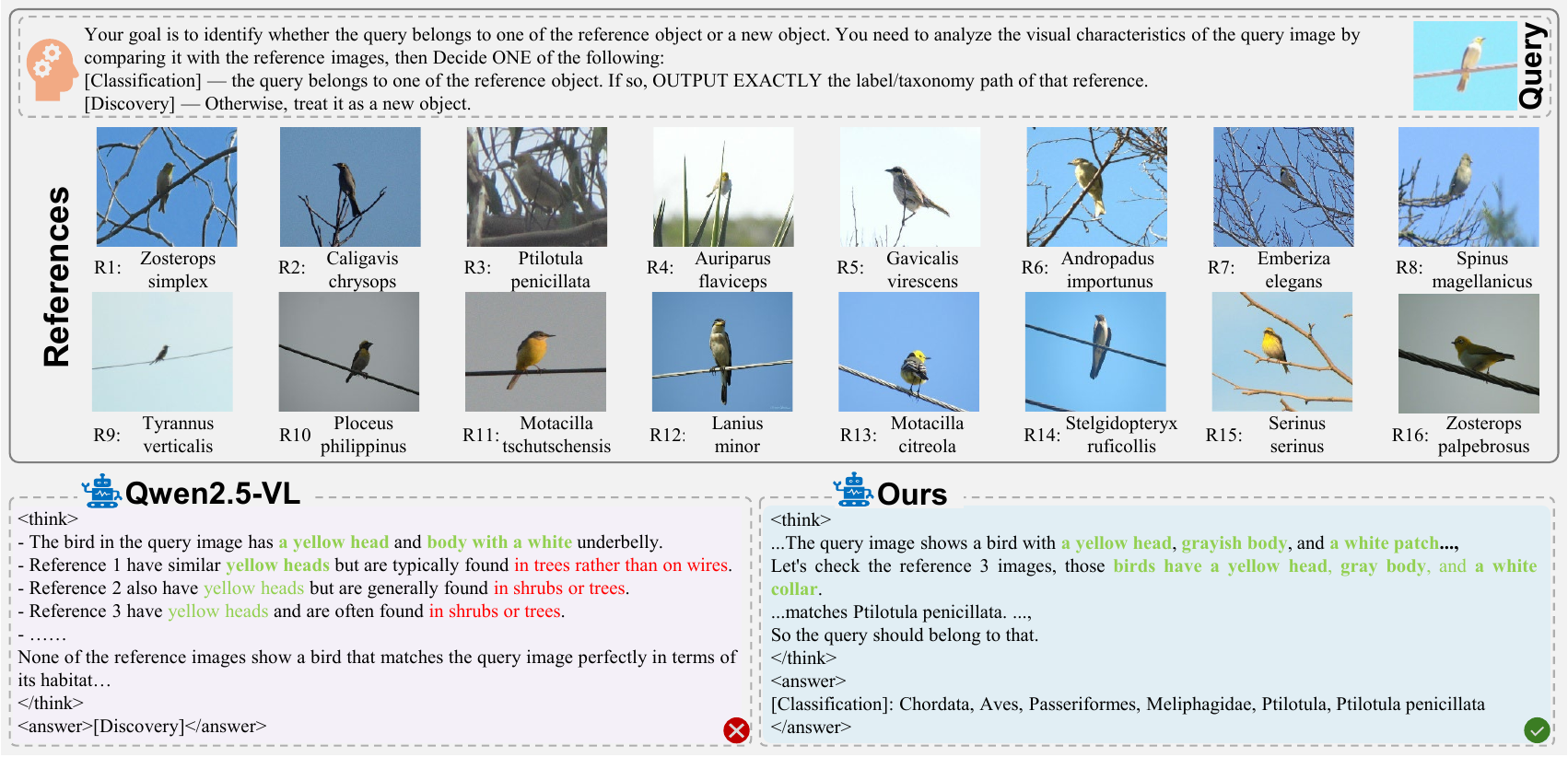}
    \caption{Bird case study: Qwen2.5-VL hallucinates habitat-based reasoning (``found in trees rather than on wires'') and outputs Discovery. DeepTaxon grounds reasoning in visual features (yellow head, gray body, white collar) and correctly classifies as \textit{Ptilotula penicillata}.}
    \label{fig:casestudy_bird}
\end{figure*}

\begin{figure*}[t]
    \centering
    \includegraphics[width=0.95\textwidth]{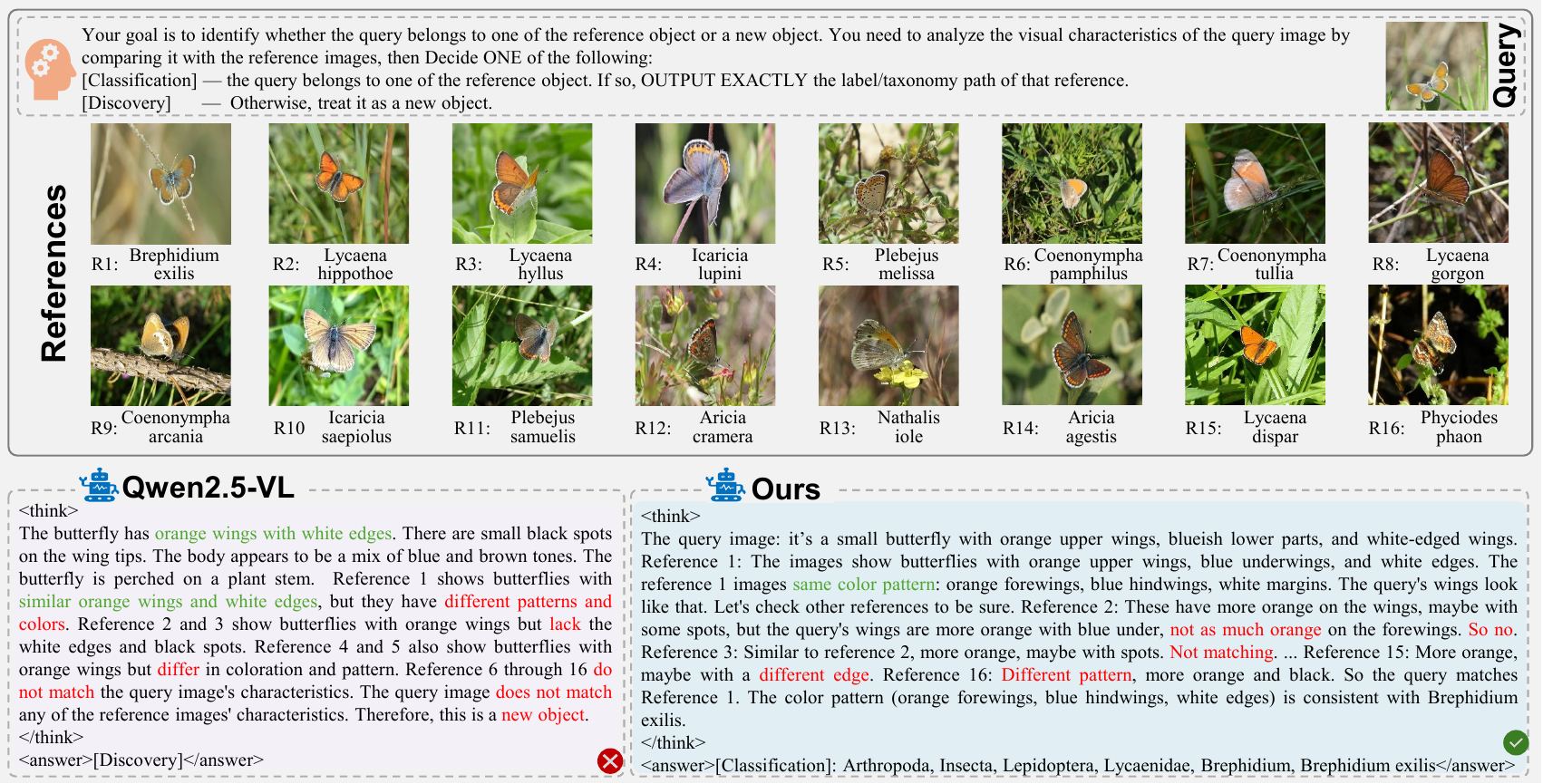}
    \caption{Butterfly case study: Qwen2.5-VL acknowledges similar features in R1 but vaguely dismisses the match. DeepTaxon systematically compares color patterns (orange forewings, blue hindwings, white margins), correctly classifying as \textit{Brephidium exilis}.}
    \label{fig:casestudy_butterfly}
\end{figure*}

\section{Case Studies}
\label{app:casestudy}

We present two qualitative examples illustrating how DeepTaxon's retrieval-augmented comparative reasoning overcomes common failure modes of zero-shot multimodal LLMs. Both examples are drawn from biologically-related domains where fine-grained discrimination is most challenging.

\textbf{Bird identification (Figure~\ref{fig:casestudy_bird}).}
The query is a \textit{Ptilotula penicillata} (White-plumed Honeyeater) photographed on a wire, with 16 retrieved candidates including the correct species at R3. Zero-shot Qwen2.5-VL correctly identifies the yellow head but is distracted by an invented ecological prior, reasoning that references with ``similar yellow heads'' are ``typically found in trees rather than on wires.'' This habitat-based reasoning is not grounded in any visual evidence from the reference images and leads to a spurious Discovery verdict. DeepTaxon instead focuses on discriminative morphological attributes: the yellow head, grayish body, and white collar pattern. It systematically compares these features against R3's reference images and outputs the correct taxonomic classification. This example illustrates how our retrieval-augmented training pipeline teaches the model to ground decisions in observed visual features rather than hallucinated ecological priors.

\textbf{Butterfly identification (Figure~\ref{fig:casestudy_butterfly}).}
The query is a \textit{Brephidium exilis} (Western Pygmy-Blue) with the correct species at R1 among 16 candidates. Zero-shot Qwen2.5-VL identifies the query's ``orange wings with white edges'' and acknowledges that R1 shows ``similar orange wings and white edges,'' yet dismisses the match by vaguely asserting ``different patterns and colors'' without specifying any concrete visual difference. This failure to articulate discriminative evidence leads it to reject all 16 candidates and output Discovery. In contrast, DeepTaxon performs systematic feature-level analysis, decomposing the query into three discriminative attributes, namely orange upper wings, blueish lower parts, and white-edged margins. For R1, it confirms matching patterns across the forewings (orange), hindwings (blue), and margins (white). It then explicitly eliminates R2--R16 with attribute-specific justifications (e.g., R2 exhibits ``more orange on the wings'' without the characteristic blue underside), ultimately classifying the query as \textit{Brephidium exilis}. This example demonstrates how structured comparative reasoning prevents premature rejection of correct matches by requiring evidence-based justification for each decision.

Both examples highlight that DeepTaxon grounds each decision in explicit visual comparisons with retrieved references, preventing hallucination and premature rejection failures common in zero-shot MLLMs. The interpretable reasoning chains further enable domain experts to trace exactly which morphological features drove each classification or discovery verdict, a transparency absent in both threshold-based OOD detection and black-box classification approaches.

\end{document}